\documentclass[a4paper]{article}

\usepackage{INTERSPEECH2022}
\usepackage{url}
\usepackage[colorlinks=true, allcolors=blue]{hyperref}
\usepackage{xcolor,colortbl}

\title{Qualitative Evaluation of Language Model Rescoring in Automatic Speech Recognition}
\name{Thibault Bañeras-Roux$^1$, Mickaël Rouvier$^2$, Jane Wottawa$^3$, Richard Dufour$^1$}
\address{
  $^1$LS2N - Nantes University (France)\\
  $^2$LIA - Avignon University (France)\\
  $^3$LIUM - Le Mans University (France)}
\email{thibault.roux@univ-nantes.fr, mickael.rouvier@univ-avignon.fr, jane.wottawa@univ-lemans.fr, richard.dufour@univ-nantes.fr}

\begin{document}

\maketitle
\begin{abstract}
Evaluating automatic speech recognition (ASR) systems is a classical but difficult and still open problem, which often boils down to focusing only on the word error rate (WER). However, this metric suffers from many limitations and does not allow an in-depth analysis of automatic transcription errors. In this paper, we propose to study and understand the impact of rescoring using language models  in ASR systems by means of several metrics often used in other natural language processing (NLP) tasks in addition to the WER. In particular, we introduce two measures related to morpho-syntactic and semantic aspects of transcribed words: 1) the POSER (Part-of-speech Error Rate), which should highlight the grammatical aspects, and 2) the EmbER (Embedding Error Rate), a measurement that modifies the WER by providing a weighting according to the semantic distance of the wrongly transcribed words. These metrics illustrate the linguistic contributions of the language models that are applied during a posterior rescoring step on transcription hypotheses.

\end{abstract}
\noindent\textbf{Index Terms}: Automatic speech recognition, Semantic analysis, Language modeling, evaluation metrics

\section{Introduction}

Over the last years, various speech and language processing fields have made significant progress thanks to scientific and technological advances. Automatic Speech Recognition (ASR)  has notably benefited from the massive increase in available data and the use of deep learning approaches~\cite{deng2013new,amodei2016deep}, making its models more robust and efficient~\cite{baevski2020wav2vec}. From an application point of view, several usage contexts are possible: an automatic transcription can either be used directly ({\it e.g.} for automatic subtitling), or it can be part (often as an input) of another application ({\it e.g.} human-computer dialogue, automatic indexing of audio documents, etc.). Despite the current performance, errors in automatic transcriptions are inevitable and impact its use: for example, ASR errors can affect applications where these systems are implemented, and thus negatively influence their global performance by making it difficult for humans to understand the transcriptions.

ASR systems are widely evaluated with the Word Error Rate (WER) metric. The simplicity of this metric is its main advantage and the reason of its massive adoption, as it only requires a reference transcription ({\it i.e.} manually annotated) of the words. It is nevertheless limited in the sense that no other information than the word itself is integrated ({\it e.g.} no linguistic information is taken into account, no semantic knowledge, etc.). Each error also has the same weight within this metric even though we know that words have a different impact considering a targeted task~\cite{morchid2016impact}. These limitations have already been exposed in the past, with proposed variants such as the IWER~\cite{mdhaffar2019qualitative}, which focuses on words chosen as {\it important} within a transcription.

In this paper, we investigate a set of automatic measures used in various natural language processing (NLP) tasks to help in the specific evaluation of ASR systems, especially on language-related aspects. These measures should allow for a finer analysis of transcription errors, by highlighting certain forms of the errors (part-of-speech, context errors, semantic distance, etc.). One of the advantages of these proposed measures is that they do not require any additional manual annotation of transcriptions and can be applied to any language. Moreover, their multiplication allows us to put forward different visions of the errors, these metrics can then complement each other. We then propose a qualitative analysis using these metrics on a state-of-the-art ASR system, by analyzing in more details the contribution of a posteriori reordering of transcription hypothesis, a process called rescoring, performed with a quadrigram language model (LM) coupled to a Recurrent Neural Net Language Model (RNNLM) on a French dataset.

This paper is organized as follows: in Section~\ref{s:prop_metric}, we describe the classical WER metric, before listing and detailing the different automatic measures we propose to allow a finer evaluation of transcriptions at a linguistic level. In order to understand the interest of these measures, a qualitative analysis of language model rescoring is proposed, first detailing the experimental protocol in Section~\ref{s:prot_expe}, then the results and analysis in Section~\ref{s:results}. Finally, a conclusion as well as perspectives are provided in Section~\ref{s:conclu}.

\section{Description of proposed measures}
\label{s:prop_metric}

ASR systems are mainly evaluated through the WER. In this section, we first describe it (Section~\ref{s:wer}) in order to highlight its advantages and limitations. Then we detail the 6 complementary automatic measures that we wish to apply to the evaluation of automatic transcriptions at the syntactic (Sections~\ref{s:cer},~\ref{s:per} and~\ref{s:lemma}) and semantic (Sections~\ref{s:eer},~\ref{s:bs} and~\ref{s:sd}) levels in addition to the WER.

\subsection{Word Error Rate (WER)}
\label{s:wer}

This metric compares a reference (manual) transcription with an automatic transcription obtained with an ASR system on the word level, words being a chain of characters between two blanks. The WER then simply takes into account three types of errors: substitutions (S), insertions (I) and deletions (D). 

\begin{itemize}
\item \textit{Substitution (S)}: in a given word chain, one transcribed word was different from the reference word.
\item \textit{Insertion (I)}: in a given word chain, a transcribed word was inserted with respect of the reference. The hypothesis counts one word more than the reference. 
\item \textit{Deletion (D)}: in a given word chain, a word in the reference was not transcribed. The hypothesis counts one word less than the reference. 
\end{itemize}

The following example sentences illustrates an alignment between a reference sentence ({\it Reference}) and an automatic transcription ({\it Hypothesis}) allowing the calculation of the WER:

\begin{table}[h!]
\centering
\scalebox{0.85}{
\begin{tabular}{c | c c c c c c c }
   {\bf Reference} & How & are & you & & today & Patrick\\
   & {\it S} & {\it D} &  = & {\it I} & = & {\it S}\\
   {\bf Hypothesis} & Were & & you & here & today & playing \\
\end{tabular}
}
\label{tabWER}
\end{table}

Formally, the WER is calculated as follows:


{\small 
\begin{equation}
WER = \frac{\# S + \# I + \# D}{\# reference\ words}
\end{equation}
}


By definition, the WER therefore considers any type of error of equivalent importance. This is the main advantage of this metric: its simplicity of application and use. However, the WER does have limitations. Using the previous example, the word {\it Patrick} was transcribed as {\it playing}. An alternative transcription hypothesis could have been {\it Patricia}. In both cases, the WER would be identical to the reference, even though the nature of the error is different ({\it Patricia} is in the same grammatical category while {\it playing} is different from the reference word in terms of syntactics and semantics). Another limitation concerns the few categories considered (substitution, insertion, deletion) for the rate calculation carrying no additional information about the context. 

\subsection{Character Error Rate (CER)}
\label{s:cer}

The character error rate (CER) is based on the same principle as the WER but applied to character chains instead of word chains. It has already been used in the ASR domain~\cite{xu2021transformer}. Initially, it is particularly suitable for character-based languages such as Chinese or Japanese. For Latin languages, and in particular French, the CER allows, among other things, to give an indication of the nature of the errors: a low CER could indicate that the ASR system tends to generate words close to the reference (and thus potentially incorporating errors related to gender, number, tense, etc.) as opposed to a high CER, with transcription assumptions that are very distant from the references.

\subsection{Part-of-speech Error Rate (POSER)}
\label{s:per}

We also chose to use a metric allowing the calculation of the error rate on the part-of-speech (POS) classes of a transcription (POSER for {\it Part-of-speech Error Rate}). POSER allows us to know if the transcribed sentences are grammatically close to the reference ones, and to better characterize substitution errors. This rate is calculated with the same formula as the WER, except that POS are taken into account instead of words which relates to metadata of the transcribed words. 

\subsection{Lemma Error Rate (LER)}
\label{s:lemma}

With a concept similar to the POSER and the WER, we did a Lemma Error Rate which consists of calculating the error rate of lemmas. We did two versions of this metric : one computing the WER and one computing the CER between the lemmas of the reference and the lemmas of the hypothesis. 

\subsection{Embeddings Error Rate (EmbER)}
\label{s:eer}

As previously exposed, the semantic aspect of a transcription is not taken into account in the WER metric. To address this, we consider a metric based on lexical word embeddings. Unlike existing metrics based on word embeddings, we aim at keeping the WER  but weighting it: a word is no longer considered in a binary way ($0$ for a good transcript and $1$ for an error), errors being weighted according to their semantic distance from the reference word. This distance is computed using the cosine similarity between the embeddings of the reference word and of the substituted transcribed word.


\subsection{BERTScore} 
\label{s:bs}

Developed for text generation~\cite{zhang2019bertscore}, this metric aims at comparing a reference word and a hypothesis with respect to semantic proximity. The first step consists in obtaining the words and sub-words ({\it tokens}) of the reference and the hypothesis thanks to the WordPiece tokenizer used by BERT~\cite{devlin2018bert}.

Then, given the sequence of contextualized embeddings of reference ($x1$,...,$xk$) and hypothesis ($\hat{x}1$,...,$\hat{x}m$), the cosine similarity is computed between each reference and hypothesis embeddings to obtain a score matrix weighted here with the inverse frequency of the document~\cite{zhang2019bertscore}.

To compute the precision, we associate each token $x$ with a token $\hat{x}$ by selecting the token bringing the highest similarity. The recall is computed by associating each $\hat{x}$ token with an $x$ token in the same way. The f-measurement score, which we use in our experiments, is computed with the recall and the precision~\cite{zhang2019bertscore}.

\subsection{Sentence Semantic Distance (SemDist)}
\label{s:sd}

While previous metrics focus on words and characters, the principle of this metric~\cite{kim2021evaluating} is to consider the complete sentence. In the ASR framework, the reference and the hypothesis are respectively transformed into their sentence embeddings using a SentenceBERT~\cite{reimers2019sentence} model, {\it i.e.} a model of sentence embeddings using the contextual word embeddings of BERT~\cite{devlin2018bert}. It is then possible to compare these vectors with the cosine similarity. Our final measure is the average of the cosine similarities between each reference's sentence embeddings and its respective hypothesis.

\section{Experimental protocol}
\label{s:prot_expe}

In this section, we present the experimental protocol set up to apply the different metrics listed in Section~\ref{s:prop_metric}. We describe the data used for our qualitative analysis of language model rescoring in Section~\ref{s:data}, the ASR system and the POS tagger in Sections~\ref{s:rap} and~\ref{s:pos} respectively. Finally, we present the embeddings used by the different metrics and the lemmatizer.

\subsection{Data}
\label{s:data}

The French datasets used to train the ASR system are ESTER 1 and 2~\cite{galliano2006corpus, galliano2009ester}, EPAC~\cite{esteve2010epac}, ETAPE~\cite{gravier2012etape}, REPERE~\cite{giraudel2012repere} and internal LIA data. Taken together, the corpora represent approximately 940 hours of audio of radio and television broadcast data. The evaluation of the systems is done on the REPERE test corpus, which is about 10 hours of audio data.

\begin{table*}[!htb]
\centering
\begin{tabular}{c|c|c|c|c|c|c|c|c|c|}
\cline{2-10}
 & \textbf{WER} & \textbf{CER} & \textbf{LER} & \textbf{LCER} & \textbf{dPOSER} & \textbf{uPOSER} & \textbf{EmbER} & \textbf{SemDist} & \textbf{BERTScore} \\ \hline
\multicolumn{1}{|c|}{\textbf{WER}} &  \multicolumn{9}{c|}{\cellcolor[gray]{.9}}\\ \cline{1-2}
\multicolumn{1}{|c|}{\textbf{CER}} & 89.34 & \multicolumn{8}{c|}{\cellcolor[gray]{.9}} \\ \cline{1-3}
\multicolumn{1}{|c|}{\textbf{LER}} & 88.08 & 88.49 & \multicolumn{7}{c|}{\cellcolor[gray]{.9}} \\ \cline{1-4}
\multicolumn{1}{|c|}{\textbf{LCER}} & 87.10 & 98.31 & 91.40 & \multicolumn{6}{c|}{\cellcolor[gray]{.9}} \\ \cline{1-5}
\multicolumn{1}{|c|}{\textbf{dPOSER}} & 92.96 & 90.02 & 92.70 & 89.51 & \multicolumn{5}{c|}{\cellcolor[gray]{.9}} \\ \cline{1-6}
\multicolumn{1}{|c|}{\textbf{uPOSER}} & 90.40 & 90.58 & 93.69 & 90.81 & 97.95 & \multicolumn{4}{c|}{\cellcolor[gray]{.9}} \\ \cline{1-7}
\multicolumn{1}{|c|}{\textbf{EmbER}} & 96.51 & 91.51 & 86.57 & 88.78 & 91.00 & 88.98 & \multicolumn{3}{c|}{\cellcolor[gray]{.9}} \\ \cline{1-8}
\multicolumn{1}{|c|}{\textbf{SemDist}} & 71.81 & 64.78 & 62.22 & 62.60 & 65.33 & 64.13 & 75.73 & \multicolumn{2}{c|}{\cellcolor[gray]{.9}} \\ \cline{1-9}
\multicolumn{1}{|c|}{\textbf{BERTScore}} & 74.63 & 74.27 & 72.60 & 73.00 & 74.09 & 74.25 & 84.51 & 63.35 & \multicolumn{1}{c|}{\cellcolor[gray]{.9}} \\ \hline
\end{tabular}
\caption{Averages of the Pearson correlations between the proposed metrics from both Base and Rescoring systems. For readability reasons, the values are multiplied by 100.} \label{t:corrs}
\end{table*}

\subsection{Automatic Speech Recognition (ASR) system}
\label{s:rap}

The ASR system is based on an existing state-of-the-art recipe\footnote{\url{https://github.com/kaldi-asr/kaldi/blob/master/egs/librispeech/s5/}} that uses the Kaldi~\cite{povey2011kaldi} toolkit. The acoustic model is a deep neural network based on the TDNNF~\cite{povey2018semi} architecture. To make the system more robust to different acoustic conditions, the audio files were randomly perturbed in speed and volume ({\it i.e.} data augmentation) during the training process.

Three language models are used. The first is a trigram model trained with SRILM~\cite{stolcke2002srilm} and used directly by the ASR system. The second is a RNNLM, a deep neural network based language model, used in an a posterior rescoring process. The network consists of three TDNN layers interspersed with two LSTM layers. Also, a quadrigram model is used during the rescoring step. The training corpus and the vocabulary used to learn the trigram model, the RNNLM model and the quadrigram model are identical. The rescoring is optional as we want to observe its impact on the different metrics.

\subsection{Tools}
\label{s:pos}

We used the POET tool\footnote{\url{https://huggingface.co/qanastek/pos-french}}\cite{labrak2022antilles}, a POS tagger for French language based on Flair~\cite{akbik2018contextual} contextual embeddings and used to automatically extract the morpho-syntactic information from words. We chose this labeler because it allows us to have both the generic classes of Universal Dependency (noun, adjective, adverb, etc.) but also a fine granularity thanks to additional information on these same labels (feminine plural noun, third person plural personal pronoun, etc.). We then propose two measures based on POS tags derived from the POSER (Section \ref{s:per}): one integrating the detailed classes (dPOSER) and one with the generic classes of Universal Dependency (uPOSER). Note that no manual POS tag annotation was used: both reference and hypothesis transcripts were automatically tagged. To obtain the lemmas, we used the Spacy lemmatizer for French\footnote{\url{https://github.com/explosion/spacy-models/releases/tag/fr_core_news_md-3.2.0}}.

For the EmbER metric (Section \ref{s:eer}), we used Fasttext embeddings~\cite{bojanowski2017enriching} and applied an error of 0.1 if the cosine similarity is above a threshold of 0.4, and 1 in other cases. The threshold was decided empirically given the cosine similarity between synonyms compare to cosine similarity between words randomly chosen.

For the SemDist metric (Section \ref{s:sd}), the multilingual SentenceBERT embeddings was used. Finally, for the BERTScore, we use the default multilingual-BERT base model\footnote{\url{https://github.com/Tiiiger/bert_score}}.


\section{Experiments and Analysis}
\label{s:results}





This section presents firstly an analysis of the six applied metrics presented in Section~\ref{s:prop_metric} in addition to the WER, and secondly a qualitative  study of the impact of the language model rescoring process used in our ASR system.

\subsection{Metrics analysis}
\label{s:ana}


\begin{table*}[!htb]
\centering
\begin{tabular}{|c||c|c||c|c|c|c||c|c|c|}
\hline
\textbf{System} & \textbf{WER} & \textbf{CER} & \textbf{dPOSER} & \textbf{uPOSER} & \textbf{LER} &\textbf{LCER} & \textbf{SemDist} & \textbf{BERTScore} & \textbf{EmbER} \\ \hline\hline
\textbf{Base} & 15.45 & 8.57 & 14.59 & 12.22 & 14.35 & 8.78 & 7.89 & 9.12 & 12.33 \\ \hline
\textbf{Rescoring} & 13.24 & 7.70 & 12.51 & 10.79 & 12.08 & 8.00 & 7.18 & 8.38 & 10.79 \\ \hline\hline
\textit{Reduction} & -14.3 \% & -10.2 \% & -14.3 \% & -11.7 \% & -15.8 \% & -8.8 \% & -9.0 \% & -8.1 \% & -12.5 \% \\\hline
\end{tabular}
\caption{Performance comparison of the Base and Rescoring systems using different metrics. The observed reduction between the two systems, in relative value, is also provided.
} \label{results}
\end{table*}

In order to make a more in-depth analysis of our metrics, in particular to understand and estimate the links that they can maintain between them, we calculated a Pearson correlation between our different measurements for our two systems and averaged them in Table~\ref{t:corrs}. 
The higher the score between two metrics, the more they are considered correlated. Clearly, the first remark is that not all metrics correlate with each other in the same way. 
SemDist is the metric that correlates the least with the others. This might be due to the fact that it is the only metric based on sentence embeddings in our experiments, going beyond the {\it word} dimension. This weak correlation implies that minimizing the WER would not correlate strongly with better performance on downstream tasks ({\it i.e.} extrinsic evaluation) using sentence embeddings. This idea is consistent with many publications in NLP and ASR that consider intrinsic ratings to be less relevant than extrinsic ratings~\cite{wang2003word,glavavs2019properly}. Indeed, the authors of SemDist~\cite{kim2021evaluating} concluded that their metric correlated better with downstream tasks than the WER.

We can see that the metric that correlates best with BERTScore and SemDist is EmbER, all three of which are based on embeddings, while the metric that correlates best with EmbER is WER. This highlights that the Embedding Error Rate is a hybrid metric that has the advantage of correlating with WER and embeddings-based metrics.

An interesting observation to make is that LER correlate the best with uPOSER and has a better correlation with dPOSER than LCER. It seems that part-of-speech and lemmas share some similarity : if the lemma is wrong, the POS is often wrong. Also, the LCER and the CER have a correlation of 0.9831 which probably means that when the CER is high, there is a good chance that the word is wrong too and so is the lemma. On the other hand, it also means that the LCER does not bring more information than the CER.

\subsection{Rescoring Impact}
\label{s:impact}



In order to improve the performance of the ASR, rescoring was acheived using a RNNLM, a deep neural network based language model. 

Table~\ref{results} 
presents the results obtained with the different metrics applied to the automatic transcriptions from the ASR system without (Base) and with hypothesis reordering (Rescoring). As expected, rescoring improves the results with a decrease of error rates independently of the used metric: an improvement is thus visible at the level of words, characters, syntax and semantics. The gains for each metric are also provided in Table~\ref{results}. They mainly highlight the fact that the relative gain obtained on the WER is the highest compared to the other metrics. Depending on the purpose of the system, the quality of a transcription can be defined by its grammatical, lexical or semantic similarity with the reference. We therefore imagine that the benefits obtained thanks to this rescoring step are not as significant as what the WER suggests. In comparison, the SemDist and BERTScore metrics have the lowest relative gains, which tends to make us say that rescoring only partially corrects transcribed words that were semantically far from their reference. The proposed EmbER, which is a mixed measure between WER and embeddings, seems to take into account the syntactic and semantic level, with a gain between that of WER and embedding measurements.  
Overall, language model rescoring contributes less to the improvement of the semantic level (SemDist, BERTScore and EmbER) compared to the syntactic level, visible with a huger reduction on the character, POS and lemma based measures.
Thanks to the meta information annotated in the REPERE corpus, we could observe that the rescoring process deteriorates performances on utterances of spontaneous speech. In average, utterances presenting more errors after the rescoring step contained 1.23 times more spontaneity information (elisions, reduction, truncations and others disfluences).



This is in line with the hypothesis we made: a speech with too much disfluences (and so a mismatch between linguistic training and testing conditions) might be negatively impacted by rescoring.

\begin{table}[!htb]
\centering
\begin{tabular}{c|c|c||c|}
\cline{2-4}
\multicolumn{1}{l|}{} & \textbf{Base} & \textbf{Rescoring} & \textbf{Reduction} \\ \hline
\multicolumn{1}{|c|}{\textbf{INTJ}} & 14.07 & 10.45 & -3.63 \\ \hline
\multicolumn{1}{|c|}{\textbf{CCONJ}} & 9.83 & 6.82 & -3.01 \\ \hline
\multicolumn{1}{|c|}{\textbf{VERB}} & 6.10 & 4.20 & -1.90 \\ \hline
\multicolumn{1}{|c|}{\textbf{ADJ}} & 5.08 & 3.41 & -1.67 \\ \hline
\multicolumn{1}{|c|}{\textbf{AUX}} & 4.66 & 3.27 & -1.39 \\ \hline
\multicolumn{1}{|c|}{\textbf{PRON}} & 5.37 & 4.12 & -1.25 \\ \hline
\multicolumn{1}{|c|}{\textbf{SCONJ}} & 3.51 & 2.43 & -1.08 \\ \hline
\multicolumn{1}{|c|}{\textbf{PROPN}} & 6.72 & 5.82 & -0.90 \\ \hline
\multicolumn{1}{|c|}{\textbf{NOUN}} & 3.34 & 2.57 & -0.77 \\ \hline
\multicolumn{1}{|c|}{\textbf{ADV}} & 3.23 & 2.49 & -0.74 \\ \hline
\multicolumn{1}{|c|}{\textbf{ADP}} & 2.90 & 2.25 & -0.65 \\ \hline
\multicolumn{1}{|c|}{\textbf{DET}} & 2.95 & 2.42 & -0.53 \\ \hline
\multicolumn{1}{|c|}{\textbf{NUM}} & 2.96 & 2.62 & -0.34 \\ \hline
\end{tabular}
\caption{Average semantic distance per POS between each word from the reference and their associated word from the hypothesis. For readability. the values are multiplied by 100.} \label{semdist:pos}
\end{table}

With respect to POS, we propose in Table~\ref{semdist:pos} to measure the average cosine distance between every reference and hypothesis word. We computed this distance without (Base) and with rescoring, while providing the relative reduction for each POS. This highlighted that Interjections (INTJ) and subordinating conjunctions (CCONJ), and to a lesser extent verbs (VERB) and adjectives (ADJ), are the word categories that benefit the most from rescoring while numbers (NUM) or determinants (DET) are among the POS classes that benefit the less from this additional step. The reason for the improvement of the interjections is probably because this POS is the one with the highest error rate. 


\section{Conclusions and Perspectives}
 \label{s:conclu}

In this study, we applied different measures in addition to the WER metric to ASR systems in order to reveal different linguistic dimensions (grammatical, semantic, etc.) to transcription errors.

We have chosen to verify their relevance by studying the impact of a posteriori hypothesis reordering on ASR systems using language models. Our study showed that the gains are not equivalent depending on the metric considered, thus highlighting the limitations of WER alone to study improvements at the lexical, grammatical or semantic level. It is important to note that the rescoring improve overall performances, though the increase in performance is not always visible locally.

In the continuity of this work, we would like to extend this analysis by combining the measures. Indeed, we have been interested here in these metrics independently, but it seems relevant to study, for example, semantic measures on identified POS ({\it e.g.}, compare BERTScore on personal names and adjectives). Also, this study focuses on the linguistic aspect of ASR, while we observed that segments with high speech spontaneity clues may be negatively impacted by the rescoring process. It would then be interesting to continue this study at the acoustic level, by looking into other audio factors such as noise or speech overlap. In the longer term, it would be interesting to evaluate the correlation between our metrics and human perception of errors.

\section{Acknowledgments}

This work was supported by the DIETS project financed by the Agence Nationale de la Recherche (ANR) under contract ANR-20-CE23-0005. It was granted access to the HPC resources of IDRIS under the allocation 2021-A0111012991 made by GENCI.

\bibliographystyle{IEEEtran}

\bibliography{mybib}


\end{document}